\documentclass[conference]{IEEEtran}
\IEEEoverridecommandlockouts
\usepackage{cite}
\usepackage{amsmath,amssymb,amsfonts}
\usepackage{algorithmic}
\usepackage{graphicx}
\usepackage{textcomp}
\usepackage{booktabs}
\usepackage{graphicx}
\usepackage{xcolor}
\usepackage{tabularx}

\begin{document}

\title{Scene and Environment Monitoring Using Aerial Imagery and Deep Learning}

\author{\IEEEauthorblockN{1\textsuperscript{st} Mahdi Maktabdar Oghaz}
\IEEEauthorblockA{\textit{dept. of Science, Eng and Computing} \\
\textit{Kingston University }\\
London, UK  \\
m.maktabdaroghaz@kingston.ac.uk}
\and
\IEEEauthorblockN{2\textsuperscript{nd} Manzoor Razaak}
\IEEEauthorblockA{\textit{dept. of Science, Eng and Computing} \\
\textit{Kingston University }\\
London, UK  \\
manzoor.razaak@kingston.ac.uk}
\and
\IEEEauthorblockN{3\textsuperscript{rd} Hamideh Kerdegari}
\IEEEauthorblockA{\textit{dept. of Science, Eng and Computing} \\
\textit{Kingston University }\\
London, UK \\
h.kerdegari@kingston.ac.uk}
\and
\IEEEauthorblockN{4\textsuperscript{th} Vasileios Argyriou}
\IEEEauthorblockA{\textit{dept. of Science, Eng and Computing} \\
\textit{Kingston University }\\
London, UK  \\
vasileios.argyriou@kingston.ac.uk}
\and
\IEEEauthorblockN{5\textsuperscript{th} Paolo Remagnino}
\IEEEauthorblockA{\textit{dept. of Science, Eng and Computing} \\
\textit{Kingston University }\\
London, UK \\
p.remagnino@kingston.ac.uk}

}

\maketitle

\begin{abstract}
Unmanned Aerial vehicles (UAV) are a promising technology for smart farming related applications. Aerial monitoring of agriculture farms with UAV enables key decision-making pertaining to crop monitoring. Advancements in deep learning techniques have further enhanced the precision and reliability of aerial imagery based analysis. The capabilities to mount various kinds of sensors (RGB, spectral cameras) on UAV allows remote crop analysis applications such as vegetation classification and segmentation, crop counting, yield monitoring and prediction, crop mapping, weed detection, disease and nutrient deficiency detection and others. A significant amount of studies are found in the literature that explores UAV for smart farming applications. In this paper, a review of studies applying deep learning on UAV imagery for smart farming is presented. Based on the application, we have classified these studies into five major groups including: vegetation identification, classification and segmentation, crop counting and yield predictions, crop mapping, weed detection and crop disease  and nutrient deficiency  detection. An in depth critical analysis of each study is provided.
\end{abstract}

\begin{IEEEkeywords}
Crop Monitoring, Image segmentation, UAV, Aerial Imagery, Deep Learning
\end{IEEEkeywords}

\section{Introduction}

The Industry 4.0 trend is transforming the production capabilities of all industries, including the agricultural domain. IoT and Artificial Intelligence are key enabling technologies for this transformation. Agriculture 4.0 will no longer depend on applying water, fertilizers, and pesticides uniformly across entire fields. Instead, farmers will use the minimum quantities required and target very specific areas. Farms and agricultural operations will have to be run very differently and more efficiently, primarily due to advancements in technology such as sensors, devices, machines, and information technology. Future agriculture will use sophisticated technologies such as robots, temperature and moisture sensors, aerial images and UAVs, behaviour and action analysis, multi-spectral and hyper-spectral imaging devices and GPS and other positioning technology. These advanced devices and precision agriculture and robotic systems will allow farms to be more profitable, efficient, safe, and environmentally friendly and paves the way toward Agriculture and Industry 4.0 \cite{Kamilaris2018, weltzien2016digital,Bloom02, Bloom03, ozdogan2017digital}.

In the past few years, the UAV industry has grown from a niche market to mainstream availability, lowering the cost of aerial imagery acquisition and opening the way to many interesting applications. This had great positive impact in various industries such as agriculture. UAVs are now increasingly used as a cost effective and timely method of capturing remote sensing images. The advantages of UAV technology include low cost, small size, safety, ecological operation, and most of all, the fast and on-demand acquisition of images.The advance of UAV technology has reached the stage of being able to provide extremely high resolution remote sensing images encompassing abundant spatial and contextual information\cite{khan2018estimation,lindner2016uav}

In recent years, aerial imagery using UAVs becomes an integral part of precision agriculture. The spatial resolution provided by UAVs, is revolutionizing precision agriculture workflows for measurement crop condition and yields over the growing season, for identifying and monitoring weeds and other applications such as registration \cite{Huang2018, Argyriou01, Argyriou02, Argyriou03}. Various studies are proposing novel applications of UAV image analysis for precision agriculture and vegetation and crop monitoring. The vast quantity of data, acquired by UAVs as well as recent advancements in parallel computing and GPU technology, enabled researcher to to adopt and deploy data driven analysis and decision making techniques such as deep learning into agriculture domain. Deep learning models which are vaguely inspired by information processing and communication patterns in biological nervous systems, have revolutionized the artificial intelligence and computer vision techniques \cite{Kamilaris2018}. Deep learning techniques are similar to ANN. However, deep learning is about “deeper” neural networks that provide a hierarchical representation of the data by means of various convolutions. This allows larger learning capabilities and thus higher performance and precision. Deep learning allow data representation in a hierarchical way, through several levels of abstraction.

A strong advantage of deep learning is automatic feature extraction from raw data, with features from higher levels of the hierarchy being formed by the composition of lower level features. Deep learning can solve more complex problems particularly well and fast, because it benefits massive parallelization.
Convolutional Neural Networks (CNN) constitute a class of deep, feed-forward ANN which mainly devised for deep operation on images. In recent years, CNN has been extensively used in computer vision and image processing. CNNs are capable to address highly complex classification and segmentation tasks that classic computer vision techniques were unable to solve \cite{lecun2015deep}.
Positive effects of deep learning technique and UAVs into agriculture domain, paved the way for agriculture 4.0 and inspired various studies and research in the last decade\cite{Kamilaris2018}.

In this regard, this study investigates the use of deep learning and UAVs imagery as two major drivers in agricultural 4.0.
Several studies have been investigating the use of  these technologies. Based on the domain and the application, we have categorized these studies into five major groups including: vegetation identification, classification and segmentation, crop counting and yield predictions, crop mapping, weed detection and crop disease  and nutrient deficiency  detection.
This study conducts a comprehensive comparison among various approaches and studies and investigates their strengths and weaknesses. The potential benefits of deep learning and UAVs in agricultural industry are not fully compromised and there are several limitations and challenges to be addressed.

The following sections present a critical analysis on the existing literature on vegetation and crop monitoring using aerial imagery and deep learning.

\section{Vegetation Identification, Classification and Segmentation}

\begin{table*}
\centering
\caption{Summary of the studies that used UAVs and deep learning for crop classification and segmentation}
\label{my-label}
\resizebox{\textwidth}{!}{%
\begin{tabular}{llll}
\hline
Author &  Crop Type & Dataset & Method \\ \hline
Ji \textit{et al.}   & Corn, forest, grass, rice, road, soybean, water, wheat & Gaofen 2 multi-temporal images & VGG CNN, 3D spatio-temporal kernel tensor \\
Chunjing \textit{et al.}  & Pond, rice, algae, waste-land, river, building, wood, road & Panchromatic images captured by  Gaofen 1 & Conventional CNN with spatio-temporal features   \\
Rebetez \textit{et al.}  & 22 different types of crops & Swiss confederation’s agroscope dataset & HistNN model using per-window histograms\\
Fan \textit{et al.} & Tobacco crop & 14 Hi-res tobacco plant images & Morphology, segmentation and CNN \\
Gao \textit{et al.} & Maize crop & In-house Lidar scans of maize field & Unsupervised clustering and faster R-CNN \\ \hline
\end{tabular}%
}
\end{table*}

Automatic classification and segmentation of crops and vegetation through UAVs is becoming a fundamental technology for vegetation identification and classification.
Over the last decade, numerous studies particularly focused on crop classification and segmentation using aerial images, computer vision and machine learning techniques. However, rise of deep learning in recent years was a breath of fresh air in this research area.
Deep learning models transform spatial, spectral, and temporal data into discriminative feature vectors and then classify each feature vector to certain types of vegetation according to the supplied training labels \cite{Chunjing2017, Shi2018}.
Many studies have moved from classic machine learning techniques to the state of the art artificial neural networks and deep learning approaches to classify and segment the aerial images of vegetation.


Ji \textit{et al.} \cite{Shi2018} proposed a three dimensional convolutional neural networks for crop classification with multi-Temporal UAV Images.
They designed a 3D kernel tensor according to the structure of multi-spectral multi-temporal remote sensing data, consisting of weights, number of channels and temporal indicator which generates a 3D feature map after 3D convolution on spatio-temporal images and accumulating over different spectral bands.
They recycled widely used neural network structure developed by Oxford’s Visual Geometry Group (VGGnet) as template to train a deep convolutional neural network where all 2D convolution operations are replaced by 3D convolution. However, the kernel size remained unchanged (3x3). All other network parameters were fine-tuned empirically for training 3D crop samples and learning spatio-temporal discriminative representations, with the full crop growth cycles being preserved.
They also introduced an active learning strategy to the CNN model to improve labeling accuracy up to a required threshold.
The use of 3D CNN is especially suitable in characterizing the dynamics of crop growth and it could outperforms over classic techniques. However the use of of the shelf VGGnet  which is not really designed to pick up the fine texture patterns in plants and vegetation could be a drawback of this study.


Chunjing \textit{et al.}\cite{Chunjing2017} investigates the application of convolutional neural network in classification of high resolution agricultural aerial images. They designed a network of 11-layer convolution neural network, including the input layer, three convolution layers, two pooling layers, two local contrast adjustment layers, two full connection layers and one output layer. Aside from the last classification layer, they employed Sigmoid activation function for all other convolutional and fully connected layers. They have taken the strong temporal and regional (Ezhou, China) characteristics of crop into account. Combined with the image feature of various types, the class category labels including: Pond, Rice, Algae, Waste-land, River, Building, Wood, Road and other planting were used in this study. The results of this study shows that convolution neural network method has significantly higher precision rate than classic machine learning techniques such as SVM.
Using sigmoid activation function increases the likelihood of vanishing gradient, also sigmoid function decreases the sparsity of the network which decreases its resolving power.


In another study Rebetez \textit{et al.}\cite{rebetez2016augmenting} proposed a  hybrid deep neural network which combines convolutional layers with per-window histograms to increase crop classification performance. A dataset of high resolution aerial images from experimental farm fields issued from a series of experiments conducted by the Swiss Confederation’s Agroscope research center used in this study.
The dataset consists of 22 different types of crops taken with RGB cameras mounted on UAV. The deep network fed with window blocks of 21x21 pixels with both RGB colors and texture features perceived as discriminative parameters for crop classification.  They proposed a deep neural network which consists of a convolutional side (CNN) which uses the raw pixel values and a dense side which uses RGB histograms (HistNN). The output of both networks was merged by a  dense layer of 128 neurons which then passed to the final layer to predict the class probability among the 22 target classes using a softmax function. This study shows the RGB histograms network outperformed the simple convolutional network in terms of classification accuracy and F-score. However the combined CNN-HistNN network generate overall superior results.
The major limitation of this study is absence of temporal information in both data collection stage and the deep model. Plants tend to have different texture and color characteristics at different time and seasons. This urges the deep learning methods to perform data acquisition at various time spans to address the color and texture diversity.


A study by Fan \textit{et al.}\cite{Fan2018} focused on automatic tobacco plant detection in UAV images via deep neural networks.
They proposed a new 3-stage algorithm based on deep neural networks to detect tobacco plants in high resolution images captured by UAVs. In the first stage, a number of candidate tobacco plant regions are extracted from UAV images with the classic computer vision approaches such as morphological operations and watershed segmentation. Each candidate region contains a tobacco plant or a non-tobacco plant to maintain the balance between two classes. In the second stage, they built a deep convolutional neural network and trained it with the purpose of classifying the candidate regions as tobacco plant regions or non-tobacco plant regions. The proposed network composed of three convolutional layers, one pooling layer and two fully connected layers. The network utilizes 3x3 convolutional kernels with stride 1 and 2x2 pooling kernels with stride 2. They employed Stochastic gradient descent as the optimizer function. In the third stage, post-processing such as Manhattan inter-class distance is performed to further remove the non-tobacco plant regions. They evaluate their model using a dataset of 14 high resolution images captured by UAVs. The experimental results show that the proposed algorithm outperformed SVM and Random forest classifiers on the detection of tobacco plants in UAV images. Deep learning techniques demand for fairly large annotated dataset for training. The annotation could be a labour intensive process. However semi-supervised approach used in this study can simplify this task and improve crop classification using deep learning techniques.


A study by Gao \textit{et al.}\cite{Gao2018} used terrestrial data and faster R-CNN deep network along with regional growth algorithms for individual maize segmentation. This study used scanned 3D points LIDAR training data and sliced them into 3D window, then points within each window were compressed into deep images. Faster R-CNN deep model has trained with these images to detect maize stem. The detected stems in the images were mapped into 3D points, which were used as seed points for the regional growth algorithm to grow individual maize from bottom to up. The results reports that their method generates promising results in individual maize segmentation.
The unsupervised maize stem clustering, waivers the labour intensive annotation process. However, the accuracy and precision of this approach is questionable.

Table 1 summarize the studies that used UAVs and deep learning for classification and segmentation of crops.

\section{Crop Counting and Yield Predictions}

\begin{table*}
\centering
\caption{Summary of the studies that used UAVs and deep learning for crop counting and yield predictions}
\label{my-label}
\resizebox{\textwidth}{!}{%
\begin{tabular}{llll}
\hline
Author &  Crop Type & Dataset & Method \\ \hline
Tri \textit{et al.}   & Paddy fields & 800 UAV images & Google inception deep sparse model \\
Rahnemoonfar \textit{et al.}  & Tomato & Synthetic images &  Modified Inception-ResNet layers   \\
Dijkstra \textit{et al.}  & Not specified & 10 sample UAV images + cell-nuclei dataset & CentroidNet model \\ \hline
\end{tabular}%
}
\end{table*}

Growth of UAV industry from a niche to mainstream market, significantly lowered the cost of aerial imagery acquisition and paved the way to many interesting applications such as crop counting and yield predictions.


Tri \textit{et al.}\cite{tri2016novel} proposed a novel approach based on deep learning techniques and UAVs for yield assessment of paddy fields. The proposed method consists of four stages including: image acquisition, image pre-processing, sampling, classifying the imagery by deep learning and yield assessment. The images were acquired by means of high resolution cameras mounted on UAV. They employed several pre-processing operations such as sliding window techniques, brightness/contrast adjustment, image rotation/flipping to the acquired images to enhanced and adopt the raw information for DNN model. They used Google inception deep sparse model to train and classify the yield value of each image. The statistic measure in rice bushes are manually collected within three small-area samples of paddy fields (about 1 square meter ). Then, they count amounts of nuts per bush, amounts of bushes per sample from which derives the yield of paddy fields per hectare.
google inception is fairly large model with mostly require transfer learning methods to perform reasonably in custom dataset. However this study attempted to train this model from the ground solely with their proprietary data which undermines its performance.


A study by Rahnemoonfar \textit{et al.}\cite{Rahnemoonfar2017} proposed real-time yield estimation based on deep learning and UAV imagery. The proposed method in this study is capable to estimates the count of tomatoes explicitly from the glance of the entire image which reduces the overhead of object detection and localization. The proposed convolutional network was trained using synthetic images and tested on real images. Rahnemoonfar \textit{et al.}\cite{Rahnemoonfar2017} claimed their approach is robust and efficient even if there is illumination variance in the images and it can also count the tomatoes which are under shadows or occluded by foliage or overlap between tomatoes which is relatively bold claim.
Their network includes a 7X7 convolution layer followed by 3X3 max pooling layer with stride of two pixels. The convolutional layers map the 3 bands (RGB) in the input image to 64 feature maps using a 7X7 kernel function.
The feature maps then fed to 2 modified Inception-ResNet layers followed the normal convolutional layers.
Inception-ResNet captures features at multiple sizes by concatenating the results of convolutional layers with different kernel sizes, and residual networks.
this feature enables the proposed model to count tomatoes with different sizes.
The experiment results claims that this method achieved over 91 percent of accuracy and less than 3 percents of residual error.
The synthetic images that used to train the network are significantly different than the real-world actual aerial images of tomato and cant be used to train a deep network. A generative adversarial networks (GAN) or an autoencoder could be a better alternative to generate larger synthetic training samples from smaller real-world set of images.


In another study, Dijkstra \textit{et al.}\cite{dijkstra2018centroidnet} proposed a deep neural network model named CentroidNet for crop localization and counting. CentroidNet relies on centroids of image objects rather than bounding boxes and combines image segmentation and centroid majority voting to regress a vector field with the same resolution as the input image. Each vector in the field points to its relative nearest centroid which makes the CentroidNet architecture independent of image size. CentroidNet can be attached to a fully convolutional networks as backbone. This study used U-Net segmentation network as a basis. A dataset of 10 frames, captured by a low-cost UAV with resolution of 3840x2160 pixels comprising crops with various sizes and heavy overlap has been created to compare CentroidNet to the other networks such as YOLOv2. Experiment results indicate that CentroidNet outperformed other detectors in detection and localization accuracy. The major drawback of this study is limited number of sample images in dataset.

Table 2 summarize the studies that used UAVs and deep learning for crop counting and yield predictions.

\section{Crop Mapping}

\begin{table*}
\centering
\caption{Summary of the studies that used UAVs and deep learning for crop mapping}
\label{my-label}
\resizebox{\textwidth}{!}{%
\begin{tabular}{llll}
\hline
Author &  Crop Type & Dataset & Method \\ \hline
Nijhawan \textit{et al.}   & Not specified & 6834 multispectral images of vegetation and 8673 images of non-vegetation area & Pre-trained Alex-net architecture \\
Baeta \textit{et al.} & Coffee & 9 Hi-res images of  images of coffee cultivation &  Multiple ConvNet models    \\ \hline
\end{tabular}%
}
\end{table*}

Crop and vegetation mapping is an important strategic technique for managing yield and agricultural products in larger scale and over an extended period of time. In recent years UAV and deep learning techniques have been adopted in various crop mapping studies.


Nijhawan \textit{et al.}\cite{Nijhawan2018} proposed a deep learning hybrid CNN framework for vegetation crop mapping. They employed an exhaustive combination of a vast number of input parameters including spectral bands, topographic and texture parameters. A new deep learning framework model that contains four individual CNNs combined together. Principle Component Analysis (PCA) used in order to find the most uncorrelated spectral, textural, and topographical information before passing them to each CNN. The proposed deep convolutional neural network architecture is based on the pre-trained Alex-net architecture which has been slightly modified to work with the data in this study. They used a network architecture consists of 8 layers, 5 out of which are  convolutional layers and the rest are fully connected ones. ReLU activation function is used for every layer which introduces non-linearity into the model. Outputs from the CNN's are then combined to form a final feature vector and classified using the SVM classifier. They used fairly large dataset, consist of 6834 images of vegetation area and 8673 for non-vegetation area which is one of the strength of this research.
The proposed method in this research is computationally expensive. A daisy chained list of exhaustive spatial and frequency domain operations on top of a relatively deep Alex-net makes this model impractical for any real-world scenario.


Baeta \textit{et al.}\cite{Baeta2017} used deep learning features on multiple scales for coffee crop recognition and mapping. They combined deep learning and fusion/selection of features from multiple scales for coffee crop recognition and mapping. Multiple ConvNet models served as deep model in this research. The proposed approach is a pixel-wise strategy that consists in the training and combination of convolutional neural networks designed to receive as input different context windows around labeled pixels. Final maps are created by combining the output of those networks for a non-labeled set of pixels. The results of this study claimed that multiple scales produces better coffee crop maps than using single scales approach. The main contribution of this study is adaptation of established scaling technique for coffee crop recognition and mapping.

Table 3 summarize the studies that used UAVs and deep learning for crop mapping.

\section{Weed Detection}

 \begin{table*}
\centering
\caption{Summary of the studies that used UAVs and deep learning for weed detection}
\label{my-label}
\resizebox{\textwidth}{!}{%
\begin{tabular}{llll}
\hline
Author &  Crop Type & Dataset & Method \\ \hline
Bah \textit{et al.}   & Weeds & 45022 labeled and 17044 unlabeled aerial images  & Background segmentation along with pretrained ResNet architecture \\
Liujun \textit{et al.}  & Morningglory, cocklebur, palmer amaranth & low-altitude and high-altitude UAV images &  Thresholding based on centroid for initial semi-supervised segmentation     \\
Ferreira \textit{et al.}  & Weeds in soybean crops & 15 thousands high resolution UAV images  &  CaffeNet   \\
Huang \textit{et al.}  & Weeds & 91 UAV images &  Modified VGG16 FCN and deconvolutional network  \\
Sa \textit{et al.}  & Weeds & 10,000 aerial UAV images  & SegNet encoder-decoder \\ \hline
\end{tabular}%
}
\end{table*}

Weeds are one of the major reasons for most agricultural yield losses. To deal with this threat, farmers resort to spraying the fields uniformly with herbicides. This method not only requires huge quantities of herbicides but impacts the environment and human health. Precision agriculture techniques allows to allocate the right doses of herbicide to the right place and at the right time which significantly reduces the costs as well as negative environmental impacts of herbicide. In recent years, UAVs are transforming to an aerial image acquisition system for weed localization and management. Despite notable advances in UAVs  acquisition systems, the automatic segmentation of weeds remains a challenging problem because of their strong similarity to the crops.
Deep learning techniques are capable to generalize complex classification and segmentation problems beyond what classic machine learning techniques were able to achieve.


Bah \textit{et al.}\cite{Bah2018a, bah2017weeds} proposed a deep learning based classification system for identifying weeds in vegetable fields such as spinach, beet and bean using high-resolution UAV imagery. They combined  deep learning with background segmentation and line detection to classify weed from actual crop. Their method comprises three main phases. First, the crop rows were automatically detected  and used to identify the inter-row weeds. In the second phase, inter-row weeds were used to constitute the training dataset of weed patches and Finally, convolutional neuronal networks used on this dataset to build a model able to detect the crop and the weeds in the images. ResNet architecture used for the classification. The proposed method is applied to high-resolution Unmanned Aerial Vehicles (UAV) images of vegetables taken about 20m above the soil. The results showed that the proposed method of weeds detection was effective in different crop fields. This study assumes that all vegetables are planted in the uniform rows and there is a distinctive borders in between each row and grown weeds. This  assumption significantly cripples the applicability of this study in the real-world.


Liujun \textit{et al.}\cite{li2016real} proposed a real-time UAV weed scout for selective weed control by adaptive robust control and machine learning techniques. They used a UAV, capable to identify weeds from far above the field and close to the canopy, and measuring the plant/weed density (weed infestation rate)/weed species. They used thresholding technique to segment the greens and basically weed canopy from the background. The crop row and their centroid line was calculated and masked by its pixel density. Then,the anomalous weed patches between the crop rows was identified and its population was mapped. Convolutional Neural Network (CNN) used for weed species classification and probability assessment. The weed distribution maps and individual weed extraction used to generate the training data labels. The preliminary result of this study shows the specific weed type could be classified using this technique. This study relies on mediocre technique such as thresholding to segment the green regions from the presumably non-green background. Thresholding technique is extremely unreliable in real-world dynamic environment.


In another study Ferreira \textit{et al.}\cite{DosSantosFerreira2017} proposed a weed detection in soybean crops using CNN. They used Dji Phantom 3 to capture over fifteen thousands high resolution images (4000x3000) of crop comprising images of the soil, soybean, broadleaf and grass weeds. Then, Simple Linear Iterative Clustering (SLIC) Superpixels algorithm used to segment the images and assist in the construction of an image dataset. Various feature extraction techniques including Gray-level co-occurrence matrix, Histogram of oriented gradients, Local binary patterns as well as color distributions used to generate the feature vector of this study. They employed CaffeNet CNN architecture which is  similar to famous AlexNet structure to classify the soybean and weed pathces. The proposed algorithm claimed to have superior accuracy compare to classic machine learning techniques such as SVM, Adaboost and Random Forest. This research put a fairly large dataset of soybean plant together which can be extreemly beneficial for research in this domain. The proposed method in this study is accurate and outperforms the classic machine learning techniques. However the computational complexity of the proposed model is extremely high for any real-time application.


Huang \textit{et al.}\cite{Huang2018a} proposed a fully convolutional network for weed mapping using unmanned aerial vehicle (UAV) imagery. A modified version of VGG16 FCN used to classify the weeds from other specimen. However, to generate the localized map, they used deconvolutional layers to recreate the segmented image. They employed transfer learning to improve generalization capability, and skip architecture was applied to increase the prediction accuracy. Performance results of FCN architecture was compared with Patch-based CNN algorithm as well as  Pixel-based CNN method. They claimed that FCN method outperformed techniques, both in terms of accuracy and efficiency. The major drawback of this study is lack of a mechanism to handle the scale variance.
In a very similar study, Huang \textit{et al.} \cite{Huang2018} proposed a semantic labeling approach for accurate weed mapping of high resolution UAV images. They adopted pretrained imageNet with residual framework in a fully convolutional form, and transferred and fine-tuned with their proprietary dataset. The ResNet and VGG-16 were used as the baseline classification architectures. They applied Atrous convolution to extend the field of view of convolutional filters. They also applied multi-scale processing to simultaneously employ several branches of Atrous convolution to feature map which expected to enhace the network capability in capturing objects at different scales. A fully connected conditional random field (CRF) was applied after the CNN to further refine the spatial details. The results claims that the proposed approach outperforms pixel-based-SVM and the classical FCN-8s.
The main drawback of this study is its extremely small training dataset. However, they have attempt to address this issue with the use of pretrained models.


Sa \textit{et al.}\cite{Sa2018} proposed a semantic weed mapping framework using aerial multispectral imaging and deep neural networks. They addressed several issues including limited ground sample distances (GSDs) in high-altitude datasets, sacrificed resolution resulting from downsampling high-fidelity images, and multispectral image alignment by adopting a stand sliding window approach that operates on only small portions of multispectral orthomosaic maps (tiles), which are channel-wise aligned and calibrated radiometrically across the entire map. To counter resolution loss, they defined the tile size to be the same as that of the DNN input. SegNet which is a popular encoder-decoder deep network used in this research. A fairly large dataset, consists of over 10,000 aerial images acquired using multispectral and RGB cameras. The results claimed that the proposed method outperformed existing approaches in both accuracy and efficiency. Use of multispectral and RGB cameras in conjuction with fairly large dataset is the main advantage of this study.

Table 4 summarize the studies that used UAVs and deep learning for weed detection.

\section{Disease and Nutrient Deficiency  Detection}

\begin{table*}
\centering
\caption{Summary of the studies that used UAVs and deep learning for disease and nutrient deficiency detection}
\label{my-label}
\resizebox{\textwidth}{!}{%
\begin{tabular}{llll}
\hline
Author &  Crop Type & Dataset & Method \\ \hline
Gennaro \textit{et al.}   & Grapevine leaf stripe disease & Hi-res multispectral images   & NDVI map made from multispectral images \\
Julio \textit{at al.}   & Late blight in potato & Multispectral aerial images  & Convolutional neural networks  \\
Poblete \textit{et al.}   & Vine water status & Multispectral images at wavelengths 530, 550, 570, 670, 700 and 800 nm  & MultiLayer Perceptron (MLP) \\
Ha \textit{et al.}    & Fusarium wilt infected radish &  139 RGB images & Local Binary Patterns and VGG-A network  \\
Hunag \textit{et al.}  & Helminthosporium Leaf Blotch  & RGB aerial images  & LeNet-5 \\ \hline

\end{tabular}%
}
\end{table*}

A significant challenge farmers continually grapple with are diseases to the crops. Early detection and diagnosis of crop diseases is crucial to reduce the damage to yield production and to further contain the disease infestation. Traditional methods of manually surveying the farms to identify infected plants and treat them is labour intensive and time consuming. Aerial surveying of large farms using satellite based technologies enables identification of infested areas, however, they are expensive in terms of time and cost, and usually cover large areas that makes it difficult to  In this context, UAV are advantageous as they can offer an aerial surveying of the farm and with the appropriate sensors mounted, disease infested crop and crop regions can be quickly identified.


Crop disease and nutrient deficiency detection via UAV employ different types of camera sensors. Multispectral cameras are popular sensors mounted on UAV for disease and nutrient deficiency identification in several studies. For instance, Gennaro \textit{et al.} \cite{di2016unmanned} presented their study using multispectral cameras on UAV for identifying grapevine leaf stripe disease. The methodology computes the NDVI map from the high-resolution images obtained from the multispectral cameras. The NDVI map allowed analyses at each plant level and compared the NDVI with the foliar symptoms of the plants to find high correlation between the indices and the grape vine disease. Their methodology relies on statistical analyses and does not consider deep learning methods for analysing the multispectral images.


The research study presented by Julio \textit{at al.} \cite{duarte2018evaluating} applied deep learning methods for prediction of severity of late blight in potato crops caused by \textit{Phytophthora infestans}. Their work used a UAV to capture images of different phenotype of potato crop in the fields with a multispectral sensor. Along with CNN, they considered other machine learning algorithms including random forests, multi layer perceptron and support vector regression. The ground truth data was generated with the help of experts who rated the severity of the infestation on the potato crop. Further, the authors exploited the spectral band differences from the multispectral images to create additional datasets with different band combinations to train the machine learning models. The results of their study showed that the random forest and the CNN models outperformed other models that were considered in the study in identifying infested potato crops from the UAV images.


Along with disease identification, spectral images of crop can be used to detect nutrient deficiencies. For instance, Poblete \textit{et al.} \cite{poblete2017artificial} demonstrated that multispectral cameras mounted UAV to predict vine water status with the help of neural network models. In their study, they carried out five flights over vineyards during two seasons to account for variability of field and plant condition. NDVI was computed from the spectral images for soil and plant classification. A MultiLayer Perceptron (MLP) neural network was applied to several different spectral bands from the images to identify the best relationship between the neural network model and the water status. ANN models of different spectral band combinations were evaluated and an accuracy range between $0.68 - 0.87$ was obtained. The results demonstrated plant stresses such as nutrient components can be evaluated from spectral bands.

The results of the studies discussed demonstrate the capabilities of multispectral and hyperspectral sensors for disease and nutrient deficiency identification. However, spectral sensors are relatively more expensive and are more complex to analyse since data from multiple spectral bands need to be combined and computed to gain an insight into the disease identification and classification. With the advancements in deep learning methods and sensor technologies, studies have demonstrated that RGB cameras can be successfully employed for crop disease identification. For instance, Ha et al. \cite{ha2017deep} proposed a CNN for detecting Fusarium wilt infected radish captured using high-resolution RGB cameras mounted on an UAV. Their system captures images of radish fields at low altitudes. The radish farm images is segmented into three regions, i.e. radish, ground and mulching film using a softmax classifier k-means clustering. A CNN is then applied for training on the segmented images and subsequent identification of health radish and Fusarian wilt of radish with 93\% accuracy.


Similarly, Hunag \textit{et al.} \cite{huang2019detection} conducted a study with a RGB camera on a UAV for identification and classification Helminthosporium leaf blotch (HLB) disease in wheat crop. The RGB camera mounted on top of a DJI Phantom4 UAV acquired images at a resolution of $4000 \times 3000$ pixels. Ground investigation consisted of generating ground truth of the disease severity into four classes: normal, light, medium, and heavy. A CNN is trained to identify the four classes and results of up to 94\% accuracy were obtained. Further, they compared the performance of the CNN with other methods such as SVM, histogram, and vegetation indices and found that CNN performs better in identifying and classifying the disease infestation.

\section{Conclusion}

Future agriculture will use sophisticated IoT technologies such as self driving agricultural machineries,temperature and moisture sensors, aerial images and UAVs, multi-spectral and hyper-spectral imaging devices and GPS and other positioning technology. The vast quantity of data, acquired by these new technologies paired with recent advancement in parallel and GPU computing, enabled researcher to to adopt and deploy data driven analysis and decision making techniques such as deep learning into agriculture domain. This advancements paved the way for resolving highly complex classification and segmentation tasks in precision agriculture. This study in particular investigated the use of deep learning and UAVs imagery as two major drivers in agricultural 4.0.

This study conducted a comprehensive comparison among various studies that used deep learning techniques and UAV based image acquisition and investigated their strengths and weaknesses. Based on the application, we categorized these studies into five major groups including: vegetation identification, classification and segmentation, crop counting and yield predictions, crop mapping, weed detection and crop disease  and nutrient deficiency  detection.
We believe the potential benefits of deep learning and UAVs in agricultural industry are not fully compromised and there are several limitations and challenges to be addressed.

\section*{Acknowledgment}

This work is co-funded by the EU-H2020 within the MONICA project under grant agreement number 732350. The Titan X Pascal used for this research was donated by NVIDIA

\bibliographystyle{IEEEtran}
\bibliography{1}

\end{document}